# Kinematic calibration of Orthoglide-type mechanisms from observation of parallel leg motions

Anatol Pashkevich[a,b], Damien Chablat[b], Philippe Wenger[b]

[a]*École des Mines de Nantes*
*4, rue Alfred-Kastler, 44307 Nantes Cedex 03, France*
*e-mail: anatol.pashkevich@emn.fr*

[b]*Institut de Recherche en Communications et Cybernétique de Nantes*
*1, rue de la Noë B.P. 6597, 44321 Nantes Cedex 3, France*
*e-mals: {Damien.Chablat, Philippe.Wenger }@irccyn.ec-nantes.fr*

**Abstract**

The paper proposes a new calibration method for parallel manipulators that allows efficient identification of the joint offsets using observations of the manipulator leg parallelism with respect to the base surface. The method employs a simple and low-cost measuring system, which evaluates deviation of the leg location during motions that are assumed to preserve the leg parallelism for the nominal values of the manipulator parameters. Using the measured deviations, the developed algorithm estimates the joint offsets that are treated as the most essential parameters to be identified. The validity of the proposed calibration method and efficiency of the developed numerical algorithms are confirmed by experimental results. The sensitivity of the measurement methods and the calibration accuracy are also studied.

*Corresponding author:   Prof. A.Pashkevich
Department of Automatics and Production Systems
École des Mines de Nantes
4, rue Alfred-Kastler BP 20722
tel.: **+ 33 (0)251 85 83 00**
fax: **+ 33 (0)251 85 83 49**
e-mail: anatol.pashkevich@emn.fr,

# 1. Introduction

Parallel kinematic machines (PKM) are commonly claimed to offer several advantages over serial manipulators, such as high structural rigidity, better payload-to-weight ratio, high dynamic capacities and high accuracy (Tlusty et al., 1999; Merlet, 2000; Wenger et al., 2001). At present, the conventional serial kinematic structures have already achieved their performance limits, which are bounded by high component stiffness required to support sequential joints, links and actuators (Tsai, 1999). Thus, the PKM are prudently considered as promising alternatives to their serial counterparts that offer faster, more flexible, less costly and more accurate solutions.

However, while the PKM usually exhibit a much better repeatability as compared to serial mechanisms, they may not necessarily posses a better accuracy, which is limited by manufacturing/assembling errors in numerous links and passive joints (Wang and Masory, 1993; Daney, 2003; Renaud et al., 2006; Fassi et al., 2007; Legnani et al., 2007). Besides, for non-Cartesian parallel architectures, some kinematic parameters (such as the encoder offsets) cannot be determined by direct measurement. These motivate intensive research on PKM calibration, which recently attracted attention of both academic and industrial experts.

Similar to the serial manipulators (Schröer et al., 1995), the PKM calibration procedures are based on the minimization of a parameter-dependent error function, which incorporates residuals of the kinematic equations (i.e. differences between the measured and computed values of the sensor readings). For the parallel manipulators, the inverse kinematic equations are considered computationally more efficient, since most PKMs admit a closed-form solution of their inverse kinematics (contrary to the direct kinematics, which is analytically solvable for the serial machines but is usually unsolvable in a closed-form for the PKM) (Innocenti, 1995; Iurascu & Park, 2003; Jeong et al., 2004; Huang et al., 2005). But the main difficulty with the inverse-kinematics-based calibration is the full-pose measurement requirement (position and orientation of the end-effector), which is very hard to implement accurately (Thomas et al., 2005). Hence, a number of studies have been directed at using the subset of the pose measurement data (Daney & Emiris, 2001), which, however, creates another problem: the identifiability of the model parameters (Besnard & Khalil, 2001).

Popular approaches in the parallel robot calibration deal with one-dimensional pose errors using a double-ball-bar system or other measuring devices (Rauf et al., 2004, 2006; Williams, 2006) as well as imposing mechanical constraints on some elements of the manipulator (Daney, 1999). However, in spite of hypothetical simplicity (joint measurements are needed only), it is hard to implement in practice since an accurate extra mechanism is required to impose these constraints. Additionally, such methods reduce the workspace size and consequently the identification efficiency (Zhuang et al., 1999).

Another category of the methods, the self- or autonomous calibration (Khalil & Besnard, 1999; Wampler et al., 1995; Zhuang, 1997; Hesselbach, 2005), is implemented by minimizing the residuals between the computed and measured values of the active and/or redundant joint sensors. Adding extra sensors at the usually unmeasured joints is very attractive from a computational point of view, since it allows getting the data in the whole workspace and potentially reduces impact of the measurement noise. However, only a partial set of the parameters may be identified in this way since the internal sensing is unable to provide sufficient information for the robot end-effector absolute location. Besides, in practice, these methods are not always economically and technologically feasible because usually it is hard to add these extra sensors to an existing mechanism.

More recently, several hybrid calibration methods were proposed that utilize *intrinsic properties* of a particular parallel machine allowing one to extract the full set of the model parameters (or the most essential of them) from a minimum set of measurements. An innovative approach was developed by Renaud et al. (2004, 2005) who applied the vision-based measurement system for the parallel manipulators calibration from the *leg observations*. In this technique, the primary data (manipulator leg poses) are extracted from the image, without any strict assumptions on the leg locations or on the corresponding end-effector poses (only leg observability is needed). While defining advantages of this method, the authors stress that the legs can be observed more easily than the end-effector and the use of a camera does not imply any modification of the mechanism. The only assumption is related to the manipulator architecture (the mechanism is actuated by linear drives located on the base). However, current accuracy of the camera-based measurements is not high enough yet to widely apply this method in industrial environment.

This paper focuses on the identification of the most essential subset of geometrical parameters (joints offsets) for the Orthoglide-type mechanisms. These mechanisms are actuated by linear drives located on the manipulator base and therefore admits technique of Renaud et al. (2004, 2005) for calibration from the leg observations. But, in contrast to the known works, our approach assumes that the leg location is observed for *specific manipulator postures*, when the tool-center-point moves along the Cartesian axes. For these postures and the nominal geometrical parameters, the legs are strictly parallel to the corresponding Cartesian planes. So, the deviation of the manipulator parameters influences on the leg parallelism that gives the source data for the parameter identification. The main advantage of this approach is the simplicity and low cost of the measuring system that can avoid using computer vision. It is composed of standard comparator indicators attached to the universal magnetic stands. It is obvious that such hardware perfectly suits industrial requirements.

The remainder of the paper is organized as follows. Section 2 describes the manipulator geometry, its inverse and direct kinematics, and also contains the sensitivity analysis of the leg parallelism at the examined postures with respect to the joint encoder offsets. Section 3 focuses on the parameter identification, with particular emphasis on the calibration accuracy under the measurement noise and selection the best set of the calibration equations. Section 4 contains experimental results that validate the proposed technique, while Section 5 summarizes the main results and contribution of the paper.

## 2. Kinematic modelling

*2.1. Manipulator geometry*

The Orthoglide is a three degrees-of-freedom parallel manipulator actuated by linear drives with mutually orthogonal axes. Its kinematic architecture is presented in Fig. 1 and includes three identical parallel chains, which will be further referred as "legs". Kinematically, each leg is formally described as *PRPaR* - chain, where *P*, *R* and *Pa* denote the prismatic, revolute, and parallelogram joints respectively (Fig.2). The output machinery (with a tool mounting flange) is connected to the legs in such a manner that the tool moves in the Cartesian x-y-z space with fixed orientation (translational motions).

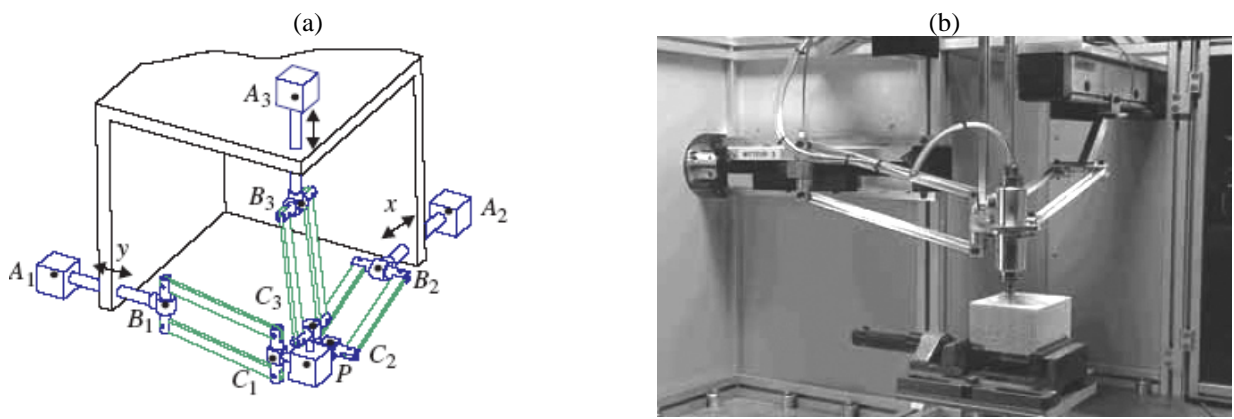

Fig. 1. The Orthoglide mechanism - kinematic architecture (a) and general view (b).

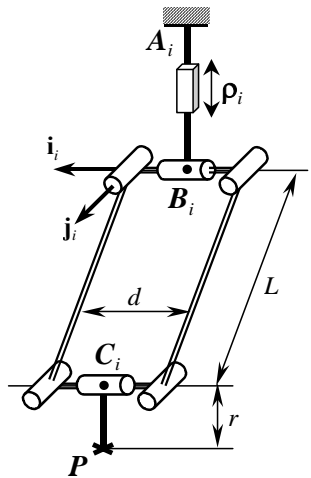

Fig 2. Kinematics of the Orthoglide leg.

In Figs. 1, 2, the base points $A_1$, $A_2$ and $A_3$ are fixed on the *i*th linear axis such that $A_1A_2 = A_1A_3 = A_1A_2$, the point $B_i$ is at the intersection of the first revolute axis $\mathbf{i}_i$ and the second revolute axis $\mathbf{j}_i$ of the *i*th parallelogram, and the point $C_i$ is at the intersection of the last two revolute joints of the *i*th parallelogram. When each $B_iC_i$ is aligned with the linear joint axis $A_iB_i$, the Orthoglide is in an *isotropic configuration* and the tool centre point $P$ is located at the intersection of the linear joint axes. In this posture, the base points $A_1$, $A_2$ and $A_3$ are equally distant from $P$. The symmetric design and the simplicity of the kinematic chains (all joints have only one degree of freedom) contribute to lower the Orthoglide manufacturing cost.

The Orthoglide is free of singularities and self-collisions. Its workspace has a regular, quasi-cubic shape. The input/output equations are simple and the velocity transmission factors are equal to one along the *x*, *y* and *z* direction at the isotropic configuration, like in a serial PPP machine (Wenger et al., 2000). The latter is an essential advantage of the Orthoglide architecture with respect to the machining applications.

Another *specific feature* of the Orthoglide mechanism, which will be further used for calibration, is displayed during the end-effector motions along the Cartesian axes. For example, for the *x*-axis motion in the Cartesian space, the sides of the *x*-leg parallelogram must also retain strictly parallel to the *x*-axis. Hence, the observed deviation of the mentioned *parallelism* may be used as the data source for the calibration algorithms.

For a small-scale Orthoglide prototype used in for the experimental part of the paper, the workspace size is approximately equal to $200 \times 200 \times 200$ mm$^3$ with the velocity transmission factors bounded between 1/2 and 2 (Chablat & Wenger, 2003). The legs nominal geometry is defined by the following parameters: $L = 310.25$ mm, $d = 80$ mm, $r = 31$ mm where $L$, $d$ are the parallelogram length and width, and $r$ is the distance between the points $\underline{C}_i$ and the tool centre point $P$ (see Fig. 2). Within the workspace, the manipulator is able to reach the Cartesian velocity of 1.2 m/s and the acceleration of 17 m/s$^2$ while carrying a payload of 4 kg.

*2.2. Modelling assumptions*

Following previous studies on the parallel mechanism accuracy (Wang & Massory, 1993; Renaud et al., 2004, Caro et al., 2006), the influence of the joint/link defects is assumed relatively small compared to the joint positioning errors that are mainly caused by the encoder offsets. The latter is also justified by the authors experience with the Orthoglide prototype, where manufacturing tolerances ±0.01 mm for the links and joints were achieved relatively easily, using common commercially available equipment. However, usual assembling techniques produced the joint offset errors about ±0.5 mm and motivated development of dedicated calibration method that are presented in this paper. These methods are based on the following modelling assumptions that are partially validated during the experimental study (see Section 4):

(i) the manipulator parts are supposed to be rigid-bodies connected by perfect joints, without clearances;

(ii) the articulated parallelograms are assumed to be identical and perfect, which insure that their sides stay parallel in pares for any motions;

(iii) the manipulator legs (composed of one prismatic joint, one parallelogram, and two revolute joints) are identical and generate a four degree-of-freedom motion each;

(iv) the linear actuator axes are mutually orthogonal and intersected in a single point to insure a translational three degree-of-freedom movement of the end-effector;

(v) The actuator encoders are assumed to be perfect but their location (zero position) is defined with some errors that are treated as the *offsets* to be estimated.

Using these assumptions, an efficient calibration technique will be developed based on the observation of the parallel motions of the manipulator legs.

*2.3. Kinematic model*

Let us first briefly present the Orthoglide kinematic model, which is described in details in the previous papers (Chablat & Wenger, 2003; Pashkevich et al., 2006).

Under the adopted assumptions, the articulated parallelograms may be replaced by kinematically equivalent single rods of the same length. Besides, a simple transformation of the Cartesian coordinates (the shift by the vector $(r, r, r)^T$) allows to eliminate the tool offset. Hence, the Orthoglide geometry can be described by a simplified model, which consists of three rigid links connected by spherical joints to the tool centre point (TCP) at one side and to the allied prismatic joints at another side (Fig. 3). Corresponding formal definition of each leg can be presented as *PSS*, where *P* and *S* denote the actuated prismatic joint and the passive spherical joint respectively.

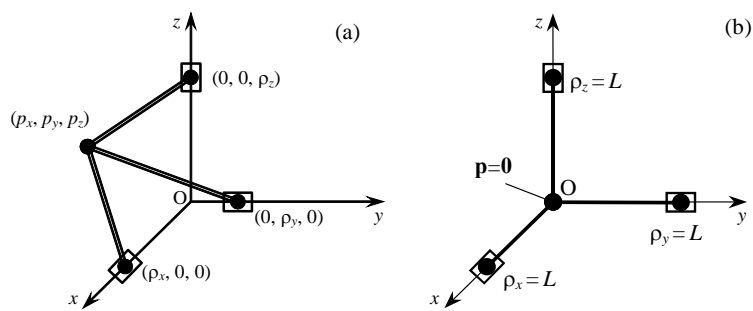

Fig. 3. Orthoglide simplified model (a) and its isotropic configuration (b).

Thus, if the origin of the reference frame is located at the intersection of the prismatic joint axes and the x, y, z-axes are directed along them, the manipulator geometry may be described by the following equations

$$\begin{aligned}
[p_x - (\rho_x + \Delta\rho_x)]^2 + p_y^2 + p_y^2 &= L^2 \\
p_x^2 + [p_y - (\rho_y + \Delta\rho_y)]^2 + p_z^2 &= L^2 \\
p_x^2 + p_y^2 + [p_z - (\rho_z + \Delta\rho_z)]^2 &= L^2
\end{aligned} \quad (1)$$

where $\mathbf{p} = (p_x, p_y, p_z)$ is the output position vector, $\boldsymbol{\rho} = (\rho_x, \rho_y, \rho_z)$ is the input vector of the prismatic joints variables, $\Delta\boldsymbol{\rho} = (\Delta\rho_x, \Delta\rho_y, \Delta\rho_z)$ is the encoder offset vector, and $L$ is the length of the parallelogram principal links. Besides, we assume that the joint variables satisfy the following prescribed joint limits

$$\rho_{\min} < \rho_i < \rho_{\max}; \quad i \in \{x, y, z\} \quad (2)$$

defined in the control software (for the Orthoglide prototype studied here, they were set as $\rho_{\min}$=-100 mm and $\rho_{\min}$=+60 mm).

It should be noted that, for this convention and for the case $\Delta\boldsymbol{\rho} = (0, 0, 0)$, the nominal isotropic posture of the manipulator corresponds to the Cartesian coordinates $\mathbf{p}_0 = (0, 0, 0)$ and to the joints variables $\boldsymbol{\rho}_0 = (L, L, L)$, see Fig. 3b. In this posture, moreover, the *x*-, and *y*-legs are oriented strictly parallel to the Cartesian plane *XY*. But the joint offsets cause the deviation of the TCP location and corresponding deviation of the parallelism, which may be computed applying the direct kinematic algorithm for the joint variables $\boldsymbol{\rho} = (L+\Delta\rho_x, L+\Delta\rho_y, L+\Delta\rho_z)$. On the other hand, in the calibration experiments, this deviation can be detected by evaluating the parallelism of the *x*- and *y*-legs with respect to the manipulator base surface (*xy*-plane). This can be easily done by measuring distances from the leg ends to the base surface and computing the difference. However, the capability of this technique is limited by evaluating the offset of the *z*-axis encoder only, since the Orthoglide mechanical design does not allow making similar measurements for the remaining pairs of the legs, with respect to the *xz*- and *yz*-planes.

Hence, within the adopted model, four parameters $(\Delta\rho_x, \Delta\rho_y, \Delta\rho_z, L)$ define the manipulator geometry, but because of the rather tough manufacturing tolerances used for the prototype, the leg link is assumed to be known and only the joint offsets $(\Delta\rho_x, \Delta\rho_y, \Delta\rho_z)$ are in the focus of the proposed calibration technique.

*2.4. Inverse and direct kinematics*

To derive calibration equations, first let us expand some previous results on the Orthoglide kinematics (Pashkevich et al., 2006) taking into account the encoder offsets. The *inverse kinematic* relations are derived from the equations (1) in a straightforward way and only slightly differ from the "nominal" case

$$\begin{aligned}
\rho_x &= p_x + s_x \sqrt{L^2 - p_y^2 - p_z^2} - \Delta\rho_x \\
\rho_y &= p_y + s_y \sqrt{L^2 - p_x^2 - p_y^2} - \Delta\rho_y \\
\rho_z &= p_z + s_z \sqrt{L^2 - p_x^2 - p_y^2} - \Delta\rho_z
\end{aligned} \quad (3)$$

where $s_x, s_y, s_z \in \{\pm 1\}$ are the configuration indices defined for the "nominal" manipulator as signs of $\rho_x - p_x$, $\rho_y - p_y$, $\rho_z - p_z$, respectively. It is obvious that expressions (3) define eight different solutions to the inverse kinematics, however the Orthoglide assembling and joint limits reduce this set for a single case corresponding to the $s_x = s_y = s_z = 1$.

For the *direct kinematics*, the equations (1) can be subtracted pair-to-pair that gives the following expression for the unknowns $p_x$, $p_y$, $p_z$ (for details, see Pashkevich et al., 2005)

$$p_i = \frac{\rho_i + \Delta\rho_i}{2} + \frac{t}{\rho_i + \Delta\rho_i}; \quad i \in \{x, y, z\} \quad (4)$$

where $t$ is an auxiliary scalar variable. This reduces the direct kinematics to the solution of a quadratic equation $At^2 + Bt + BC = 0$ with coefficients

$$A = (\rho_x + \Delta\rho_x)(\rho_y + \Delta\rho_y) + (\rho_x + \Delta\rho_x)(\rho_z + \Delta\rho_z) + (\rho_y + \Delta\rho_y)(\rho_z + \Delta\rho_z);$$
$$B = (\rho_x + \Delta\rho_x)^2(\rho_y + \Delta\rho_y)^2(\rho_z + \Delta\rho_z)^2; \quad C = \left((\rho_x + \Delta\rho_x)^2 + (\rho_y + \Delta\rho_y)^2 + (\rho_z + \Delta\rho_z)^2 - 4L^2\right)/4.$$

Of the two possible solutions $t = (-B + m\sqrt{B^2 - 4ABC})/(2A)$, $m = \pm 1$ of the quadratic formula, only the one corresponding to $m = +1$ is admitted by the orthoglide prototype (because of the selected assembly mode).

*2.5. Sensitivity analysis*

To evaluate the encoder offset influence on the legs parallelism with respect to the Cartesian planes *XY*, *YZ*, and *YZ*, let us derive first the differential relations for the TCP deviation for three types of the Orthoglide postures:

(i)  "*maximum displacement*" postures for the directions *x*, *y*, *z*  (Fig. 4a);

(ii) *isotropic* posture in the middle of the workspace (Fig. 4b);

(iii) "*minimum displacement*" postures for the directions *x*, *y*, *z* (Fig. 4c);

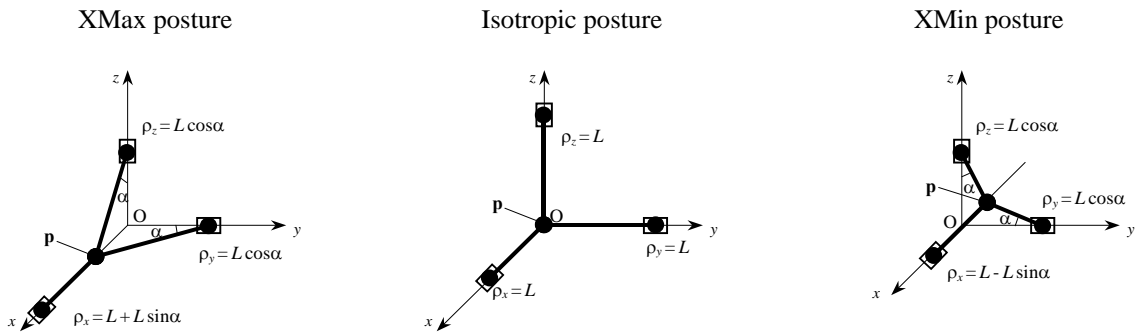

Fig. 4. Specific postures of the Orthoglide manipulator
(corresponding to the x-leg leg motion along the Cartesian axis *X* )

These postures are of particular interest for the calibration since in the "nominal" case (zero encoder offsets) the corresponding leg is parallel to the relevant pair of the Cartesian planes. On the other hand, the considered parallelism can be perturbed by the deviation of the TCP that defines location of points $C_i$ (see Fig. 2), while the opposite sides of the legs are mechanically constrained by the actuator joint axes (points $B_i$ in Fig.2).

The differential kinematical model may be derived from the Orthoglide Jacobian, the inverse of which is obtained from (1) in a straightforward way (see Pashkevich et al., 2006 for details):

$$\mathbf{J}^{-1}(\mathbf{p}, \boldsymbol{\rho}) = \frac{\partial \boldsymbol{\rho}}{\partial \mathbf{p}} = \begin{bmatrix} 1 & \dfrac{p_y}{p_x - \rho_x} & \dfrac{p_z}{p_x - \rho_x} \\ \dfrac{p_x}{p_y - \rho_y} & 1 & \dfrac{p_z}{p_y - \rho_y} \\ \dfrac{p_x}{p_z - \rho_z} & \dfrac{p_y}{p_z - \rho_z} & 1 \end{bmatrix} \qquad (5)$$

It should be noted that, for computing convenience, the above expression includes both the Cartesian coordinates $p_x, p_y, p_z$ and the joint coordinates $\rho_x, \rho_y, \rho_z$, but only one of these sets may be treated as independent because of the inverse/direct kinematic relations.

For the *isotropic posture*, the differential relations are computed in the neighbourhood of the point

$$\mathbf{p}_0 = (0, 0, 0) \quad \text{and} \quad \boldsymbol{\rho}_0 = (L, L, L),$$

which after substitution to (5) gives the identity Jacobian matrix

$$\mathbf{J}(\mathbf{p}_0, \boldsymbol{\rho}_0) = \mathbf{I}_{3\times 3} \qquad (6)$$

It means that in this case the TCP displacement is related to the joint offsets by trivial equations

$$\Delta p_i = \Delta \rho_i, \quad i \in \{x, y, z\}, \qquad (7)$$

and each joint offset influences on the TCP deviation independently and with the scaling factor of 1.0 . Taking into account the Orthoglide geometry, this deviation may be estimated by evaluating parallelism of the legs with respect to the Cartesian planes (i.e. measuring difference of distances from the leg ends to the relevant plane). However, as mentioned in subsection 2.3, this technique is feasible for the *z*-direction only, hence it may produce an estimation of $\Delta \rho_z$ merely.

For the *"maximum displacement"* posture in the *x*-direction (see Fig. 4a), the differential relations are derived in the neighbourhood of the point

$$\mathbf{p} = (L \sin \alpha, 0, 0); \quad \boldsymbol{\rho} = (L + L \sin \alpha, \ L \cos \alpha, \ L \cos \alpha)$$

where $\alpha$ is the angle between the *y*-, *z*-legs and corresponding Cartesian axes: $\alpha = \mathrm{asin}(\rho_{\max} / L)$. After the substitution into (5), this gives the inverse Jacobian as a lower triangle matrix, which admits analytical inverse yielding

$$\mathbf{J}(\mathbf{p}(\alpha), \boldsymbol{\rho}(\alpha)) = \begin{bmatrix} 1 & 0 & 0 \\ T_\alpha & 1 & 0 \\ T_\alpha & 0 & 1 \end{bmatrix}, \quad (8)$$

where $T_\alpha = \tan(\alpha)$. Hence, the differential equations for the TCP displacement may be written as

$$\Delta p_x = \Delta \rho_x; \quad \Delta p_y = T_\alpha \Delta \rho_x + \Delta \rho_y; \quad \Delta p_z = T_\alpha \Delta \rho_x + \Delta \rho_z \quad (9)$$

and the joint offset influences on the TCP deviation is estimated by factors 1.0 and $T_\alpha$. It is also worth mentioning that measurement of the *x*-leg parallelism with respect to the *XY*-plane gives an equation for estimating the offset $\Delta \rho_x$ (provided that the offset $\Delta \rho_z$ has been obtained from the isotropic posture).

Similar results are valid for the *"maximum displacement"* postures in the *y*- and *z*-directions (differing by the indices only), and also for the *"minimum displacement"* postures. In the latter case, the angle α should be computed from an equation $\alpha = \mathrm{asin}(\rho_{\min} / L)$.

Table 1.

Sensitivity of the TCP location for the representative Orthoglide postures

| Posture | Leg | Plane | Deviation | Typical value[*] |
|---|---|---|---|---|
| *Isotropic* | X | XY | $\Delta \rho_z$ | 1.00 |
| | | XZ | $\Delta \rho_y$ | 1.00 |
| | Y | XY | $\Delta \rho_z$ | 1.00 |
| | | YZ | $\Delta \rho_x$ | 1.00 |
| | Z | XZ | $\Delta \rho_y$ | 1.00 |
| | | YZ | $\Delta \rho_x$ | 1.00 |
| *Max / Min X-displacement* | X | XY | $T_\alpha \Delta \rho_x + \Delta \rho_z$ | 1.00±0.34 |
| | | XZ | $T_\alpha \Delta \rho_x + \Delta \rho_y$ | 1.00±0.34 |
| *Max / Min Y-displacement* | Y | XY | $T_\alpha \Delta \rho_y + \Delta \rho_z$ | 1.00±0.34 |
| | | YZ | $T_\alpha \Delta \rho_y + \Delta \rho_x$ | 1.00±0.34 |
| *Max / Min Z-displacement* | Z | XZ | $T_\alpha \Delta \rho_z + \Delta \rho_y$ | 1.00±0.34 |
| | | YZ | $T_\alpha \Delta \rho_z + \Delta \rho_x$ | 1.00±0.34 |

The results on the TCP sensitivity with respect to the joint offsets are summarized in Table 1 that gives also numerical values corresponding to the hypothetical joint offset $\Delta \boldsymbol{\rho} = (1 \text{ mm}, 1 \text{ mm}, 1 \text{ mm})$ and to the angle $\alpha = \pm 20°$ that are typical for the Orthoglide prototype studied in the experimental part of the paper. Analysis of these values allows concluding that the leg parallelism is rather sensitive to the joint offsets. Thus, relevant deviations $\Delta p_x$, $\Delta p_y$, $\Delta p_z$, may be used for the offset identification.

# 3. Calibration methods

*3.1 Measurement techniques*

To identify the Orthoglide kinematic parameters specified in the previous section, we propose two calibration methods, which employ different measurement techniques for the leg/surface parallelism. The first of them (Fig. 5a) assumes two measurements for the same leg posture (to assess distances from both leg ends to the base surface). The second technique assumes a fixed location of the measuring device but two distinct leg postures, which ensure positioning of the leg ends in the neighbourhood of the device. It is obvious that, for the perfectly calibrated manipulator, both methods give zero differences for each measurement pair. Conversely, the non-zero differences contain source information for the joint offset identification.

The following sub-sections contain detailed descriptions of these measurement techniques and relevant identification procedures. In particular, sub-sections 3.2 and 3.3 introduce respectively the single- and double-pose methods along with corresponding literalised calibration equations. Sub-section 3.4 describes a non-linear calibration routine that is based on the minimisation of the residual-square sum. Finally, sub-section 3.5 focuses on the calibration accuracy and sensitivity to the measurement noise.

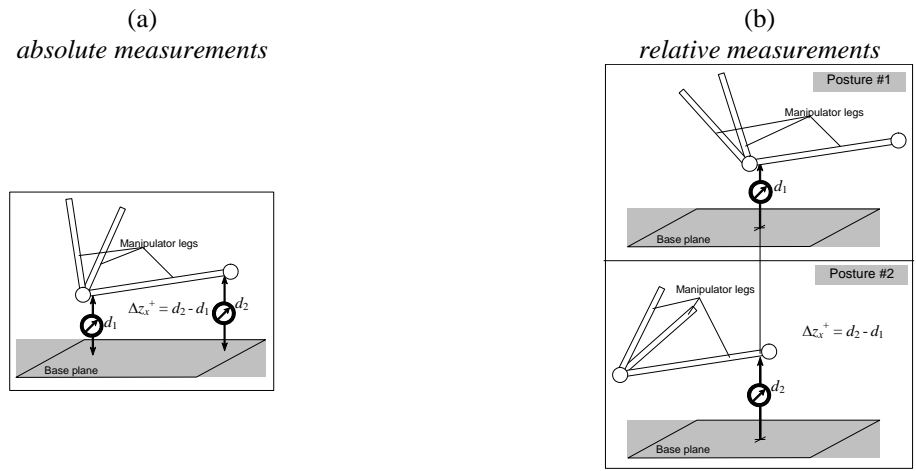

Fig. 5. Measuring the leg/surface parallelism using *single-posture-double-sensor* (a) and *double-posture-single-sensor* (b) methods.

*3.2. Calibration using single-posture measurements*

Using the single-posture measurements and taking into account the Orthoglide design limitations allowing locating gauges on the XY surface only (i.e. for the z-direction measurements), the calibration experiment may be arranged in the following way.

Step 1.  Locate the manipulator in the *isotropic* posture and measure parallelism of the *X*- and *Y*-legs with respect to the *XY*-surface: $\Delta z_x^0$, $\Delta z_y^0$

    Step 2.     Locate sequentially the manipulator in the "*X-maximum*" and "*X-minimum*" postures and measure parallelism of the *X*- legs with respect to the *XY*-surface: $\Delta z_x^+$, $\Delta z_x^-$

    Step 3.     Locate sequentially the manipulator in the "*Y-maximum*" and "*Y-minimum*" postures and measure parallelism of the *Y*- legs with respect to the *XY*-surface: $\Delta z_y^+$, $\Delta z_y^-$

In the above description, the variable following the $\Delta$-symbol denotes the measurement direction (*z* in all cases), the subscript defines the manipulator leg, and the superscript indicates the manipulator posture for this leg. For example, $\Delta z_x^+$ denotes the z-direction deviation of the X-leg for the "*X-maximum*" posture.

Using expressions from sub-section 2.5 presented in Table 1, the system of the calibration equations may be written as follows

$$\begin{bmatrix} 0 & 0 & 1 \\ 0 & 0 & 1 \\ a_1 & 0 & 1 \\ a_2 & 0 & 1 \\ 0 & a_1 & 1 \\ 0 & a_2 & 1 \end{bmatrix} \cdot \begin{bmatrix} \Delta \rho_x \\ \Delta \rho_y \\ \Delta \rho_z \end{bmatrix} = \begin{bmatrix} \Delta z_x^0 \\ \Delta z_y^0 \\ \Delta z_x^+ \\ \Delta z_x^- \\ \Delta z_y^+ \\ \Delta z_y^- \end{bmatrix} \quad (10)$$

where $a_1 = T_{\alpha 1}$ and $a_2 = T_{\alpha 2}$, which may be also computed as $a_1 = \rho_{max}/\sqrt{L^2 - \rho_{max}^2}$ and $a_2 = \rho_{min}/\sqrt{L^2 - \rho_{min}^2}$. For instance, for the Orthoglide prototype (see subsection 2.1) $a_1 \approx 0.20$ and $a_2 \approx -0.34$.

This overdetermined system of six linear equations in three unknowns may be solved in a straightforward way, using the Moore-Penrose pseudoinverse. However, from the application point of view, it is worth to separate the equations for three pairs and sequentially solve them for $\Delta \rho_x, \Delta \rho_y, \Delta \rho_z$ : this approach yields the following expressions for the joint offsets

$$\begin{aligned} \Delta \rho_z &= \frac{\Delta z_x^0 + \Delta z_y^0}{2} \\ \Delta \rho_x &= \frac{a_1(\Delta z_x^+ - \Delta \rho_z) + a_2(\Delta z_x^- - \Delta \rho_z)}{a_1^2 + a_2^2} \\ \Delta \rho_y &= \frac{a_1(\Delta z_y^+ - \Delta \rho_z) + a_2(\Delta z_y^- - \Delta \rho_z)}{a_1^2 + a_2^2} \end{aligned} \quad (11)$$

which are computationally convenient but may produce slightly higher residuals than the standard pseudoinverse.

However, the measurement procedure for this method is rather complicated in comparison with an alternative one, described in the following subsection. It should be stressed that the single-posture method requires separate measurements of $d_1$ and $d_2$ (see Fig. 5a) that are further used for computing the difference $d_2 - d_1$, while the alternative technique directly evaluates this difference using a single measuring device. It is obvious that the first method is based on the absolute measurements that are very sensitive to the gauge calibration, while the second approach (based on the relative measurements) does not require any calibration of the gauges.

*3.3. Calibration using double-posture measurements*

Since in this case a single gauge is used only, it is possible to assess the leg parallelism with respect to both relevant planes (XY and XZ for the X-leg, for instance). This advantage is charged however by using two legs postures, allowing sequentially locating both leg ends close to the gauge. For this measuring technique, the calibration experiment may be arranged in the following way:

Step 1. Locate the manipulator in the *isotropic* posture and place two gauges in the middle of the X-leg ensuring required measurement directions (orthogonal to the leg and parallel to the Cartesian axes Y and Z); get the gauge readings.

Step 2. Locate sequentially the manipulator in the "*X-maximum*" and "*X-minimum*" postures, get the gauge readings, and compute differences $\Delta y_x^+$, $\Delta z_x^+$, $\Delta y_x^-$, $\Delta z_x^-$

Step 3+. Repeat steps 1, 2 for the Y- and Z-legs and compute differences $\Delta x_y^+$, $\Delta z_y^+$, $\Delta x_y^-$, $\Delta z_y^-$, and $\Delta x_z^+$, $\Delta y_z^+$, $\Delta x_z^-$, $\Delta y_z^-$.

The system of calibration equations can be also derived using expressions from Table 1, but in two steps. First, it is required to define the gauge location that is assumed to be positioned at the leg middle point in the *isotropic* posture.[*] Hence, for the X-leg for instance, it is the midpoint of the line segment bounded by the TCP ($\Delta\rho_x$, $\Delta\rho_y$, $\Delta\rho_z$) and the centre of the X-axis prismatic joint ($L+\Delta\rho_x$, 0, 0). This yields the following differential expressions for the leg midpoints:

$$\begin{array}{ll} \text{X-leg Gauges:} & (L/2+\Delta\rho_x;\quad \Delta\rho_y/2\ ;\quad \Delta\rho_z/2\ ) \\ \text{Y-leg Gauges:} & (\ \Delta\rho_x/2\ ;\quad L/2+\Delta\rho_x;\quad \Delta\rho_z/2\ ) \\ \text{Y-leg Gauges:} & (\ \Delta\rho_x/2\ ;\quad \Delta\rho_y/2\ ;\quad L/2+\Delta\rho_z) \end{array}$$

Afterwards, in the "*X-maximum*" posture, the X-leg location is also defined by two points, namely, (i) the TCP, and (ii) the centre of the X-axis prismatic joint. Their coordinates are defined as follows (see Fig. 4a and Table 1)

$$\begin{array}{ll} \text{Tool centre point:} & (LS_\alpha+\Delta\rho_x\ ;\quad T_\alpha\Delta\rho_x+\Delta\rho_y;\quad T_\alpha\Delta\rho_x+\Delta\rho_z) \\ \text{X-joint centre:} & (L+LS_\alpha+\Delta\rho_x\ ;\quad 0\ ;\quad 0\quad ) \end{array}$$

Then, the equations of a straight-line passing along the X-leg may be written as

$$\begin{aligned} x &= \mu\,(LS_\alpha+\Delta\rho_x)+(1-\mu)\,(L+LS_\alpha+\Delta\rho_x) \\ y &= \mu\,(T_\alpha\Delta\rho_x+\Delta\rho_y);\quad z=\mu\,(T_\alpha\Delta\rho_x+\Delta\rho_z) \end{aligned} \quad (12)$$

---

[*] This assumption is not critical here because, as follows from relevant analysis, potential errors in the initial location of the gauge produce identification errors that are negligible as compared to the measurement noise.

where $S_\alpha = \sin(\alpha)$; $T_\alpha = \tan(\alpha)$, and µ is a scalar parameter, $\mu \in [0, 1]$. Since the gauge x-coordinate remains the same independently of the current posture, the parameter µ may be obtained from the equation $x = L/2 + \Delta\rho_x$, which gives the following solution:

$$\mu = 0.5 + S_\alpha. \tag{13}$$

Hence, the Y- and Z-gauge readings for the X-leg in the "X-maximum" posture are

$$\begin{aligned} y &= (0.5 + S_\alpha) T_\alpha \, \Delta\rho_x + (0.5 + S_\alpha) \, \Delta\rho_y \\ z &= (0.5 + S_\alpha) T_\alpha \, \Delta\rho_x + (0.5 + S_\alpha) \, \Delta\rho_z \end{aligned} \tag{14}$$

and, finally, the deviations of the X-leg measurements while it changes its posture from the "X-maximum" to the *isotropic* one are

$$\begin{aligned} \Delta y_x^+ &= (0.5 + S_\alpha) T_\alpha \, \Delta\rho_x + S_\alpha \, \Delta\rho_y \\ \Delta z_x^+ &= (0.5 + S_\alpha) T_\alpha \, \Delta\rho_x + S_\alpha \, \Delta\rho_z \end{aligned} \tag{15}$$

A similar approach may be applied to the "X-minimum" posture, as well as to the equivalent postures for the Y- and Z-legs. This gives the following system of twelve linear equations in three unknowns

$$\begin{bmatrix} b_1 & c_1 & 0 \\ c_1 & b_1 & 0 \\ b_2 & c_2 & 0 \\ c_2 & b_2 & 0 \\ \hline 0 & b_1 & c_1 \\ 0 & c_1 & b_1 \\ 0 & b_2 & c_2 \\ 0 & c_2 & b_2 \\ \hline b_1 & 0 & c_1 \\ c_1 & 0 & b_1 \\ b_2 & 0 & c_2 \\ c_2 & 0 & b_2 \end{bmatrix} \cdot \begin{bmatrix} \Delta\rho_x \\ \Delta\rho_y \\ \Delta\rho_z \end{bmatrix} = \begin{bmatrix} \Delta x_y^+ \\ \Delta y_x^+ \\ \Delta x_y^- \\ \Delta y_x^- \\ \Delta y_z^+ \\ \Delta z_y^+ \\ \Delta y_z^- \\ \Delta z_y^- \\ \Delta x_z^+ \\ \Delta z_x^+ \\ \Delta x_z^- \\ \Delta z_x^- \end{bmatrix} \tag{16}$$

where $b_i = \sin\alpha_i$; $c_i = (0.5 + \sin\alpha_i)\tan\alpha_i$ and $\alpha_1 = \operatorname{asin}(\rho_{\max}/L) > 0$; $\alpha_2 = \operatorname{asin}(\rho_{\min}/L) < 0$. For instance, for the Orthoglide prototype (see subsection 2.1) $b_1 \approx 0.19$, $c_1 \approx 0.14$ and $b_2 \approx -0.32$, $c_2 \approx 0.06$.

The reduced version of this system may be obtained if one assesses the leg/plane parallelism by the difference between the "maximum" and "minimum" postures. The latter leads to the system of six linear equations in three unknowns

$$\begin{bmatrix} b & c & 0 \\ c & b & 0 \\ 0 & b & c \\ 0 & c & b \\ b & 0 & c \\ c & 0 & b \end{bmatrix} \begin{bmatrix} \Delta \rho_x \\ \Delta \rho_y \\ \Delta \rho_z \end{bmatrix} = \begin{bmatrix} \Delta x_y \\ \Delta y_x \\ \Delta y_z \\ \Delta z_y \\ \Delta x_z \\ \Delta z_x \end{bmatrix} \qquad (17)$$

where $b = b_1 - b_2$; $c = c_1 - c_2$ and $\Delta x_y = \Delta x_y^+ - \Delta x_y^-$; $\Delta y_x = \Delta y_x^+ - \Delta y_x^-$, etc. For the Orthoglide prototype this values are as follows: $b \approx 0.52$, $c \approx 0.20$.

Both systems (16) and (17) may be solved using the pseudoinverse of Moore-Penrose, which ensures minimizing the residual square sum. But as follows from the simulation study, for rather essential joint offsets (about 5 mm and more) the differential equations may produce non-accurate results. For this reason, the next subsection focuses on the non-linear calibration equations and their solution through the straightforward minimization of the square sum of the residuals.

*3.4. Non-linear calibration equations*

From a general point of view, the considered calibration problem may be presented as the fitting of the experimental data to the Orthoglide kinematic model incorporating the joint offsets. Hence, it is necessary to obtain numerical algorithms that allow computing all the examined deviations for any given offsets.

To present relevant results in a concise form, let us introduce special notations for the direct and inverse kinematic models of the "nominal" Orthoglide (with zero offsets):

$$\mathbf{p} = f_0(\mathbf{\rho}); \quad \mathbf{\rho} = f_0^{-1}(\mathbf{p}) \qquad if \qquad \Delta \mathbf{\rho} = \mathbf{0} \qquad (18)$$

Then, in the *isotropic* posture, the TCP position may be expressed as

$$\left[ p_x^0, p_y^0, p_z^0 \right] = f_0 \left( L + \Delta \rho_x, L + \Delta \rho_y, L + \Delta \rho_z \right), \qquad (19)$$

while expressions for the position of the prismatic joints remain the same:

$$\begin{aligned} \text{X-leg Prismatic Joint:} & \quad (L + \Delta \rho_x \quad 0 \quad 0 \quad ) \\ \text{Y-leg Prismatic Joint:} & \quad (0 \quad L + \Delta \rho_y \quad 0 \quad ) \\ \text{Y-leg Prismatic Joint:} & \quad (0 \quad 0 \quad L + \Delta \rho_z) \end{aligned}$$

Hence, the leg midpoints defining the gauge locations may be computed as follows:

$$\begin{aligned} x_x^g &= L/2 + (p_x^0 + \Delta \rho_x)/2; & y_x^g &= p_y^0/2; & z_x^g &= p_z^0/2; \\ x_y^g &= p_x^0/2; & y_y^g &= L/2 + (p_y^0 + \Delta \rho_y)/2; & z_y^g &= p_z^0/2; \\ x_z^g &= p_x^0/2; & y_z^g &= p_y^0/2; & z_z^g &= L/2 + (p_z^0 + \Delta \rho_z)/2; \end{aligned} \qquad (20)$$

where the subscripts 'x, y, z' define the leg and the subscript 'g' refers to the gauge.

For the "X-maximum" posture, the TCP position is computed as

$$\left[ p_x^{+x}, p_y^{+x}, p_z^{+x} \right] = f_0 \left( L + LS_\alpha + \Delta\rho_x, LC_\alpha + \Delta\rho_y, LC_\alpha + \Delta\rho_z \right), \tag{21}$$

where $C_\alpha = \cos(\alpha);\ S_\alpha = \sin(\alpha)$, while the position of the X-link prismatic joints is described by the expression $(L + LS_\alpha + \Delta\rho_x;\ 0;\ 0)$. Hence, the equations of a straight-line passing along the X-leg may be written as

$$\begin{aligned} x &= \mu\, p_x^{+x} + (1-\mu)(L + LS_\alpha + \Delta\rho_x) \\ y &= \mu\, p_y^{+x}; \qquad z = \mu\, p_z^{+x} \end{aligned} \tag{22}$$

where μ is a scalar parameter, as above, which is determined by the x-coordinate of the gauge $x = x_x^g$. Solution of this equation yields

$$\mu^{+x} = \frac{L/2 + LS_\alpha - p_x^0/2 + \Delta\rho_x/2}{L + LS_\alpha - p_x^{+x} + \Delta\rho_x} \tag{23}$$

that allows one to compute the Y- and Z-gauge readings for the X-leg as $\mu^{+x} \cdot p_y^{+x}$ and $\mu^{+x} \cdot p_z^{+x}$ respectively and to get the final expression for the desired deviations of the X-leg:

$$\begin{aligned} \delta y_x^+ (\Delta\boldsymbol{\rho}) &= \mu^{+x} \cdot p_y^{+x} - p_y^0/2 \\ \delta z_x^+ (\Delta\boldsymbol{\rho}) &= \mu^{+x} \cdot p_z^{+x} - p_z^0/2 \end{aligned} \tag{24}$$

where symbol δ(.) is used to distinguish functions of the joint offsets Δρ and the experimental values, which are denoted by Δ.

A similar approach may be applied to the "X-minimum" posture, as well as to the equivalent postures for the Y- and Z-legs. Relevant expressions are summarized in Table 2 where symbol '±' stands for both the "maximum" and "minimum" postures and angle α is defined by the joint limits: $\alpha_1 = \mathrm{asin}(\rho_{max}/L) > 0$; $\alpha_2 = \mathrm{asin}(\rho_{min}/L) < 0$.

The obtained expressions allow posing the following optimisation problem for the joint offset identification

$$F = \left( \delta x_y^+ (\Delta\boldsymbol{\rho}) - \Delta x_y^+ \right)^2 + \left( \delta x_z^+ (\Delta\boldsymbol{\rho}) - \Delta x_z^+ \right)^2 + \mathrm{K} \to \min_{\Delta\boldsymbol{\rho}}, \tag{25}$$

which gives the desired values of $\Delta\rho_x$, $\Delta\rho_y$, $\Delta\rho_z$. It may be also presented in the reduced form by replacing the pairs of the deviations $(\Delta x_y^+, \Delta x_y^-)$, $(\Delta y_x^+, \Delta y_x^-)$, etc. by their differences $\Delta x_y = \Delta x_y^+ - \Delta x_y^-$; $\Delta y_x = \Delta y_x^+ - \Delta y_x^-$, etc. Both problems may be solved numerically by means of the standard gradient search technique using the Jacobians from Eqs. 16 and 17.

Table 2
Expressions for the non-linear calibration model

| Content | Expressions |
|---|---|
| TCP locations | $[p_x^0, p_y^0, p_z^0] = f_0(L + \Delta\rho_x,\ L + \Delta\rho_y,\ L + \Delta\rho_z)$ <br> $[p_x^{\pm x}, p_y^{\pm x}, p_z^{\pm x}] = f_0(L \pm LS_\alpha + \Delta\rho_x,\ LC_\alpha + \Delta\rho_y,\ LC_\alpha + \Delta\rho_z)$ <br> $[p_x^{\pm y}, p_y^{\pm y}, p_z^{\pm y}] = f_0(LC_\alpha + \Delta\rho_x,\ L \pm LS_\alpha + \Delta\rho_y,\ LC_\alpha + \Delta\rho_z)$ <br> $[p_x^{\pm z}, p_y^{\pm z}, p_z^{\pm z}] = f_0(LC_\alpha + \Delta\rho_x,\ LC_\alpha + \Delta\rho_y,\ L \pm LS_\alpha + \Delta\rho_z)$ |
| Scaling factors | $\mu^{\pm x} = \dfrac{L/2 \pm LS_\alpha - p_x^0/2 + \Delta\rho_x/2}{L \pm LS_\alpha - p_x^{\pm x} + \Delta\rho_x}$ <br><br> $\mu^{\pm y} = \dfrac{L/2 \pm LS_\alpha - p_y^0/2 + \Delta\rho_y/2}{L \pm LS_\alpha - p_y^{\pm y} + \Delta\rho_y}$ <br><br> $\mu^{\pm z} = \dfrac{L/2 \pm LS_\alpha - p_z^0/2 + \Delta\rho_z/2}{L \pm LS_\alpha - p_x^{\pm z} + \Delta\rho_z}$ |
| Leg deviations | $\delta y_x^\pm(\Delta\boldsymbol{\rho}) = \mu^{\pm x} \cdot p_y^{\pm x} - p_y^0/2;\quad \delta z_x^\pm(\Delta\boldsymbol{\rho}) = \mu^{\pm x} \cdot p_z^{\pm x} - p_z^0/2;$ <br> $\delta x_y^\pm(\Delta\boldsymbol{\rho}) = \mu^{\pm y} \cdot p_x^{\pm y} - p_x^0/2;\quad \delta z_y^\pm(\Delta\boldsymbol{\rho}) = \mu^{\pm y} \cdot p_z^{\pm y} - p_z^0/2;$ <br> $\delta x_z^\pm(\Delta\boldsymbol{\rho}) = \mu^{\pm z} \cdot p_x^{\pm z} - p_x^0/2;\quad \delta y_z^\pm(\Delta\boldsymbol{\rho}) = \mu^{\pm z} \cdot p_y^{\pm z} - p_y^0/2;$ |

*3.5. Calibration accuracy*

Because of the measurement noise, the developed technique may produce the biased estimates of the model parameters. Thus, for practical application, it is worth to evaluate the statistical properties of the calibration errors.

Within the linear calibration equations, the impact of the measurement noise may be evaluated using general techniques from the identification theory, under the standard assumptions concerning the measurement errors $\xi_i$: zero-mean independent and identically distributed Gaussian random variables with the standard deviation σ. Let us consider separately two cases corresponding to the six-equation and twelve-equation systems (7), (8), since they differ in residual covariance.

For both linear systems (16) and (17), the variance-covariance matrix of the identification parameters is written as (Ljung, 1999)

$$\mathbf{V}(\Delta\boldsymbol{\rho}) = (\mathbf{J}^T\mathbf{J})^{-1} \cdot \mathbf{J}^T \cdot \mathbf{E}(\Delta\mathbf{s} \cdot \Delta\mathbf{s}^T) \cdot \mathbf{J} \cdot (\mathbf{J}^T\mathbf{J})^{-1} \qquad (26)$$

where $\mathbf{E}(.)$ denotes the mathematical expectation, $\mathbf{J}$ is the Jacobian, and $\Delta\mathbf{s}$ is the vector of the measurement errors.

In the *six-equation case*, the vector $\Delta s$ consists of the statistically independent components corresponding to the deviations $\Delta x_y, \Delta y_x, \text{K } \Delta y_z$ and is expressed through differences of the measurement errors at the min/max leg postures:

$$\Delta \mathbf{s}_{(6)} = \left[ \xi_x^{+y} - \xi_x^{-y}, \; \xi_y^{+x} - \xi_y^{-x}, \; \text{K}, \; \xi_z^{+x} - \xi_z^{-x} \right]^T. \tag{27}$$

where the subscripts and the superscripts are defined similar to subsection 3.4. Hence, the covariance is the 6×6 identity matrix

$$\mathbf{E}\left( \Delta \mathbf{s}_{(6)} \cdot \Delta \mathbf{s}_{(6)}^T \right) = 2\sigma^2 \cdot I_{6 \times 6} \tag{28}$$

and the expression (26) is reduced to

$$\mathbf{V}(\Delta \boldsymbol{\rho}) = 2(\mathbf{J}_{(6)}^T \mathbf{J}_{(6)})^{-1} \cdot \sigma^2 \tag{29}$$

However, in the *twelve-equation case*, the vector $\Delta s$ includes some dependent components

$$\Delta \mathbf{s}_{(6)} = \left[ \xi_x^{+y} - \xi_x^{0x}, \; \xi_y^{+x} - \xi_y^{0x}, \; \xi_x^{-y} - \xi_x^{0x}, \; \xi_y^{-x} - \xi_y^{0x} \text{ K}, \; \xi_z^{-x} - \xi_z^{0z} \right]^T, \tag{30}$$

corresponding to the pairs $(\Delta x_y^+, \Delta x_y^-)$, $(\Delta y_x^+, \Delta y_x^-) \ldots (\Delta z_x^+, \Delta z_x^-)$, since each leg deviations are measured twice (for the Max/Min postures) but with respect to the same isotropic location. So, the covariance is the 12×12 non-identity matrix

$$\mathbf{E}\left( \Delta \mathbf{s}_{(12)} \cdot \Delta \mathbf{s}_{(12)}^T \right) = \sigma^2 \cdot \mathbf{G}_{12 \times 12} \tag{31}$$

expressed as

$$\mathbf{G}_{12 \times 12} = \begin{bmatrix} \mathbf{G}_{4 \times 4} & 0 & 0 \\ 0 & \mathbf{G}_{4 \times 4} & 0 \\ 0 & 0 & \mathbf{G}_{4 \times 4} \end{bmatrix}; \quad \mathbf{G}_{4 \times 4} = \begin{bmatrix} 2 & 0 & 1 & 0 \\ 0 & 2 & 0 & 1 \\ 1 & 0 & 2 & 0 \\ 0 & 1 & 0 & 2 \end{bmatrix}.$$

Consequently, the covariance (26) is presented as

$$\mathbf{V}(\Delta \boldsymbol{\rho}) = (\mathbf{J}_{(12)}^T \mathbf{J}_{(12)})^{-1} \cdot \mathbf{J}_{(12)}^T \mathbf{G} \, \mathbf{J}_{(12)} \cdot (\mathbf{J}_{(12)}^T \mathbf{J}_{(12)})^{-1} \cdot \sigma^2 \tag{32}$$

These expressions allow us to compute a scalar performance measure for the calibration accuracy $\sigma_\rho$ that may be defined as the square-averaged standard deviation of the calibration errors for the joint offsets $\Delta \rho_x, \Delta \rho_y, \Delta \rho_z$

$$\sigma_\rho = \sqrt{\frac{1}{3} trace\left( \mathbf{V}(\Delta \boldsymbol{\rho}) \right)} \tag{33}$$

where the subscript 'ρ' is used for distinguishing with the standard deviation of the measurement noise $\sigma$.

For the Orthoglide prototype described in subsection 2.1, the latter expression yields $\sigma_\rho \approx 2.06 \cdot \sigma$ in the case of twelve equations and $\sigma_\rho \approx 1.98 \cdot \sigma$ in the six-equation case. This justifies using the six-equation method because of simplicity and slightly higher identification accuracy in comparison with the twelve-equation technique.

While confirming this conclusion theoretically, it is worth mentioning that reduction of the equation number from 12 to 6 usually increases the calibration error by the factor $\sqrt{2}$. However, using the deviations $\Delta x_y, \Delta y_x, \mathrm{K}\, \Delta y_z$ (measured between the Max and Min postures) instead of $\Delta x_y^+, \Delta x_y^-, \mathrm{K}\, \Delta y_z^-$ (measured between the isotropic and Max/Min postures) increases the deviation measurement sensitivity that gives reduction of $trace((\mathbf{J}^T\mathbf{J})^{-1})$. In particular, for the case study, $|\rho_{\max}/L| \approx 0.19$ and $|\rho_{\min}/L| \approx 0.32$ while $(\rho_{\max} - \rho_{\min})/L \approx 0.52$. It means that the sensitivity increase compensates reduction of the equation number.

For the non-linear calibration equations (see subsection 3.4), the impact of the measurement errors was investigated using the Monte-Carlo method. The simulation results (obtained for 20 replications with 10000 runs for $\sigma = 0.01$ mm and two values of $\Delta\rho$) are presented in Table 3. They coincide with the above linear-approximation expressions and also justify advantages of the six-equation method for the practical applications.

Table 3
Simulation results on impact of the measurement errors for $\sigma = 0.01$ mm

| Calibration technique | std($\Delta\rho$) (offset 0.1 mm) | std($\Delta\rho$) (offset 1.0 mm) |
|---|---|---|
| *Six-equation method* | 0.0198 mm (±0.0003) | 0.0199 mm (±0.0002) |
| *Twelve-equation method* | 0.0207 mm (±0.0003) | 0.0207 mm (±0.0004) |

## 4. Experimental results

*4.1. Experimental setup*

The measuring system is composed of standard comparator indicators attached to the universal magnetic stands allowing fixing them on the manipulator bases. The indicators have a resolution of 10 μm and are sequentially used for measuring the X-, Y-, and Z-leg parallelism while the manipulator moves between the Max, Min and isotropic postures (it is obvious that for industrial applications, it is better to use more sophisticated, high precision digital indicators with the resolution of 1 μm or less, which yield more accurate calibration results).

For each measurement, the indicators are located on the mechanism base in such a manner that a corresponding leg is admissible for the gauge contact for all intermediate posters (Fig. 6). The Min and Max postures are constrained by the software joint limits and defined as $\rho_{min}$=-100 mm and $\rho_{max}$= 60 mm respectively. The initial position of the indicator corresponds to the leg middle for the manipulator isotropic posture.

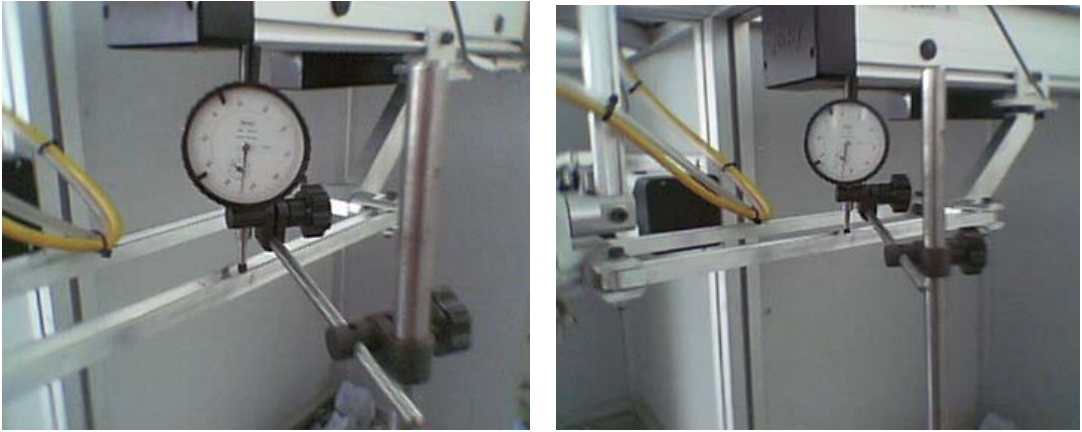

Fig. 6. Experimental Setup.

During experiments, the legs were moved sequentially via the following postures: Isotropic → Max → Min → Isotropic → … . To reduce the measurement errors, the measurements were repeated three times for each leg. Then, the results were averaged and used for the parameter identification. It should be noted that the measurements demonstrated very high repeatability compared to the encoder resolution (dissimilarity was less than 0.02 mm).

*4.2. Calibration results and their analysis*

To validate the developed calibration technique and the adopted modelling assumptions, we carried out three experiments targeted to the following objectives:

*Experiment* #1: validation of modelling assumptions (it lead to the mechanical retuning )

*Experiment* #2: collecting experimental data used for the parameter identification;

*Experiment* #3: validation of calibration results using the identified model parameters.

*Experiment* **#1**. The first calibration experiment produced rather high parallelism deviation, up to 2.37 mm as shown in Table 4. It was unexpected since the Orthoglide demonstrated quite good quality and accuracy of milling in previous tests. However, the milling tests were perfect just because of the high uniformity of the Orhoglide workspace due to the advantages of the manipulator architecture.

The straightforward application of the proposed calibration algorithm to this data set was not optimistic: in the frames of the adopted kinematic model, the root-mean-square (r.m.s.) deviation for the legs can be reduced down from 1.19 mm to 1.07 mm only (see Table 4). On the other hand, the statistical estimation of the measurement noise parameter σ (based on the residual analysis) also yielded an unrealistic result: σ ≈ 1.0 mm. It impels to conclude that the manipulator mechanics requires more careful tuning, especially location of the linear actuator axes that are assumed to be mutually orthogonal and intersected in a single point (see subsection 2.2). Thus, the manipulator mechanics was retuned, in particular the locations of the actuator axes were adjusted mechanically using the single-pose measurement technique described in subsection 3.2.

*Experiment* **#2**. The second calibration experiment (after mechanical tuning) yielded lower parallelism deviations, less than 0.70 mm (see Table 4), which is on average twice better than in the first experiment. For these data, the developed calibration algorithm yielded the joint offsets that are expected to reduce the root-mean-square deviation down from 0.62 mm to 0.28 mm, i.e. by three times. Besides, the estimated value of $\sigma \approx 0.28$ mm is more realistic taking into account both the measurement accuracy and the manufacturing/assembling tolerances. Accordingly, the identified values of the joint offsets $\Delta\rho_x = -0.53$ mm, $\Delta\rho_y = 0.59$ mm, $\Delta\rho_y = -1.76$ mm were incorporated in the Orthoglide control software.

*Experiment* **#3**. The third experiment was targeted to the validation of the calibration results, i.e. assessing the leg parallelism while using the model parameters identified from the second data set. It demonstrated very good agreement with the expected values of $\Delta x_y$, $\Delta x_z$, …$\Delta z_y$. In particular, the maximum deviation reduced down to 0.34 mm (expected 0.28 mm), and the root-mean-square value decreased down to 0.21 mm (expected 0.20 mm).

On the other hand, further adjusting of the kinematic model to the third data set gives both negligible improvement of the deviations and very small alteration of the model parameters (see Tables 4 and 5). It is evident that further reduction of the parallelism deviation is bounded by the manufacturing and assembling errors or, probably, the non-geometric errors.

*Discussion*. As follows from the above analysis, the calibration experiments confirm validity of the proposed identification technique and its ability to tune the joint offsets from observations of the leg parallelism. The achieved accuracy coincides with the quality of the Orthoglide prototype manufacturing and assembling.

Another related conclusion deals with the comparison of the six-equation and twelve-equation identification methods (see subsections 3.4 and 3.5) using real data sets, which do not necessary follow the classical assumptions on the measurement errors (Gaussian zero-mean random variables). As follows from Table 5, both methods produced roughly the same values of the model parameters, however the six-equation method is more computationally attractive and, thus, more suitable for the practice.

Table 4
Experimental data and expected improvements of accuracy

| Data Source | $\Delta x_y$ mm | $\Delta x_z$ mm | $\Delta y_x$ mm | $\Delta y_z$ mm | $\Delta z_x$ mm | $\Delta z_y$ mm | r.m.s. mm |
|---|---|---|---|---|---|---|---|
| *Initial settings (before mechanical tuning and calibration)* | | | | | | | |
| Experiment #1 | +0.52 | +1.58 | +2.37 | -0.25 | -0.57 | -0.04 | 1.19 |
| Expected improvement | -0.94 | +0.63 | +1.07 | -0.84 | -0.27 | +0.35 | 0.74 |
| *After mechanical tuning (before calibration)* | | | | | | | |
| Experiment #2 | -0.43 | -0.37 | +0.42 | -0.18 | -1.14 | -0.70 | 0.62 |
| Expected improvement | -0.28 | +0.25 | +0.21 | -0.14 | -0.13 | +0.09 | 0.20 |
| *After calibration* | | | | | | | |
| Experiment #3 | -0.23 | +0.27 | +0.34 | -0.10 | -0.09 | +0.11 | 0.21 |
| Expected improvement | -0.29 | +0.23 | +0.25 | -0.17 | -0.10 | +0.08 | 0.20 |

Table 5
Model parameters obtained using the six- and twelve equation methods

| Experiment | Calibration method | Model parameters | | | Residual r.m.s. mm |
|---|---|---|---|---|---|
| | | $\Delta\rho_x$ mm | $\Delta\rho_x$ mm | $\Delta\rho_x$ mm | |
| Experiment #1 | *Six-equation* | 2.17 | 1.69 | -1.42 | 0.74 |
| | *Twelve-equation* | 2.07 | 1.66 | -1.30 | 0.75 |
| Experiment #2 | *Six-equation* | -0.53 | 0.59 | -1.76 | 0.20 |
| | *Twelve-equation* | -0.52 | 0.55 | -1.69 | 0.21 |
| Experiment #3 | *Six-equation* | 0.07 | 0.14 | 0.00 | 0.20 |
| | *Twelve-equation* | 0.12 | 0.00 | 0.10 | 0.21 |

# 5. Conclusions

Recent advances in parallel robot architectures encourage related research on kinematic calibration of parallel mechanisms. This paper proposes a new calibration method for parallel manipulators, which allows efficient identification of the joint offsets using observations of the manipulator leg parallelism with respect to the base surface. Presented for the Orthoglide-type mechanisms, this approach may be also applied to other manipulator architectures that admit parallel leg motions (along the Cartesian axes) or, in more general cases, that allow locating the leg in several postures with a common intersection point.

The proposed calibration technique employs a simple and low-cost measuring system composed of standard comparator indicators attached to the universal magnetic stands. They are sequentially used for measuring the deviation of the relevant leg location while the manipulator moves the tool-centre-point in the directions $x$, $y$ and $z$. From the measured differences, the calibration algorithm estimates the joint offsets that are treated as the most essential parameters that are difficult to identify by other methods.

The presented theoretical derivations deal with the sensitivity analysis of the proposed measurement method, selecting the best set of the calibration equation, and also with the calibration accuracy. It has been proved that the highest accuracy is achieved for the measuring the leg parallelism at the extreme leg postures, while additional measurements at the isotropic posture does not reduce the identification error. The validity of the proposed approach and the efficiency of the developed numerical algorithm were confirmed by the calibration experiments with the Orthoglide prototype, which allowed reducing the residual root-mean-square by three times.

To increase the calibration precision, future work will focus on the development of the specific assembling fixture ensuring proper location of the linear actuators and also on the expanding the set of the identified model parameters and compensation of the non-geometric errors that are not compensated within the frames of the adopted model.

# Figure captions

Fig. 1. The Orthoglide mechanism - kinematic architecture (a) and general view (b).

Fig 2. Kinematics of the Orthoglide leg.

Fig. 3. Orthoglide simplified model (a) and its isotropic configuration (b).

Fig. 4. Specific postures of the Orthoglide manipulator corresponding to the x-leg leg motion along the Cartesian axis *X*

Fig. 5. Measuring the leg/surface parallelism using *single-posture-double-sensor* (a) and *double-posture-single-sensor* (b) methods.

Fig. 6. Experimental Setup.

Table 1
Sensitivity of the TCP location for the representative Orthoglide postures

| Posture | Leg | Plane | Deviation | Typical value[*] |
|---|---|---|---|---|
| *Isotropic* | X | XY | $\Delta\rho_z$ | 1.00 |
| | | XZ | $\Delta\rho_y$ | 1.00 |
| | Y | XY | $\Delta\rho_z$ | 1.00 |
| | | YZ | $\Delta\rho_x$ | 1.00 |
| | Z | XZ | $\Delta\rho_y$ | 1.00 |
| | | YZ | $\Delta\rho_x$ | 1.00 |
| *Max / Min X-displacement* | X | XY | $T_\alpha \Delta\rho_x + \Delta\rho_z$ | 1.00±0.34 |
| | | XZ | $T_\alpha \Delta\rho_x + \Delta\rho_y$ | 1.00±0.34 |
| *Max / Min Y-displacement* | Y | XY | $T_\alpha \Delta\rho_y + \Delta\rho_z$ | 1.00±0.34 |
| | | YZ | $T_\alpha \Delta\rho_y + \Delta\rho_x$ | 1.00±0.34 |
| *Max / Min Z-displacement* | Z | XZ | $T_\alpha \Delta\rho_z + \Delta\rho_y$ | 1.00±0.34 |
| | | YZ | $T_\alpha \Delta\rho_z + \Delta\rho_x$ | 1.00±0.34 |

Table 2
Expressions for the non-linear calibration model

| Content | Expressions |
|---|---|
| TCP locations | $\left[p_x^0, p_y^0, p_z^0\right] = f_0\left(L + \Delta\rho_x,\ L + \Delta\rho_y,\ L + \Delta\rho_z\right)$ <br><br> $\left[p_x^{\pm x}, p_y^{\pm x}, p_z^{\pm x}\right] = f_0\left(L \pm LS_\alpha + \Delta\rho_x,\ LC_\alpha + \Delta\rho_y,\ LC_\alpha + \Delta\rho_z\right)$ <br><br> $\left[p_x^{\pm y}, p_y^{\pm y}, p_z^{\pm y}\right] = f_0\left(LC_\alpha + \Delta\rho_x,\ L \pm LS_\alpha + \Delta\rho_y,\ LC_\alpha + \Delta\rho_z\right)$ <br><br> $\left[p_x^{\pm z}, p_y^{\pm z}, p_z^{\pm z}\right] = f_0\left(LC_\alpha + \Delta\rho_x,\ LC_\alpha + \Delta\rho_y,\ L \pm LS_\alpha + \Delta\rho_z\right)$ |
| Scaling factors | $\mu^{\pm x} = \dfrac{L/2 \pm LS_\alpha - p_x^0/2 + \Delta\rho_x/2}{L \pm LS_\alpha - p_x^{\pm x} + \Delta\rho_x}$ <br><br> $\mu^{\pm y} = \dfrac{L/2 \pm LS_\alpha - p_y^0/2 + \Delta\rho_y/2}{L \pm LS_\alpha - p_y^{\pm y} + \Delta\rho_y}$ <br><br> $\mu^{\pm z} = \dfrac{L/2 \pm LS_\alpha - p_z^0/2 + \Delta\rho_z/2}{L \pm LS_\alpha - p_x^{\pm z} + \Delta\rho_z}$ |
| Leg deviations | $\delta y_x^\pm(\Delta\boldsymbol{\rho}) = \mu^{\pm x} \cdot p_y^{\pm x} - p_y^0/2;\quad \delta z_x^\pm(\Delta\boldsymbol{\rho}) = \mu^{\pm x} \cdot p_z^{\pm x} - p_z^0/2;$ <br> $\delta x_y^\pm(\Delta\boldsymbol{\rho}) = \mu^{\pm y} \cdot p_x^{\pm y} - p_x^0/2;\quad \delta z_y^\pm(\Delta\boldsymbol{\rho}) = \mu^{\pm y} \cdot p_z^{\pm y} - p_z^0/2;$ <br> $\delta x_z^\pm(\Delta\boldsymbol{\rho}) = \mu^{\pm z} \cdot p_x^{\pm z} - p_x^0/2;\quad \delta y_z^\pm(\Delta\boldsymbol{\rho}) = \mu^{\pm z} \cdot p_y^{\pm z} - p_y^0/2;$ |

Table 3

Simulation results on impact of the measurement errors for $\sigma = 0.01$ mm

| Calibration technique | std($\Delta\rho$) (offset 0.1 mm) | std($\Delta\rho$) (offset 1.0 mm) |
|---|---|---|
| *Six-equation method* | 0.0198 mm ($\pm 0.0003$) | 0.0199 mm ($\pm 0.0002$) |
| *Twelve-equation method* | 0.0207 mm ($\pm 0.0003$) | 0.0207 mm ($\pm 0.0004$) |

Table 4
Experimental data and expected improvements of accuracy

| Data Source | $\Delta x_y$ mm | $\Delta x_z$ mm | $\Delta y_x$ mm | $\Delta y_z$ mm | $\Delta z_x$ mm | $\Delta z_y$ mm | r.m.s. mm |
|---|---|---|---|---|---|---|---|
| *Initial settings* (*before mechanical tuning and calibration*) | | | | | | | |
| Experiment #1 | +0.52 | +1.58 | +2.37 | -0.25 | -0.57 | -0.04 | 1.19 |
| Expected improvement | -0.94 | +0.63 | +1.07 | -0.84 | -0.27 | +0.35 | 0.74 |
| *After mechanical tuning* (*before calibration*) | | | | | | | |
| Experiment #2 | -0.43 | -0.37 | +0.42 | -0.18 | -1.14 | -0.70 | 0.62 |
| Expected improvement | -0.28 | +0.25 | +0.21 | -0.14 | -0.13 | +0.09 | 0.20 |
| *After calibration* | | | | | | | |
| Experiment #3 | -0.23 | +0.27 | +0.34 | -0.10 | -0.09 | +0.11 | 0.21 |
| Expected improvement | -0.29 | +0.23 | +0.25 | -0.17 | -0.10 | +0.08 | 0.20 |

Table 5
Model parameters obtained using the six- and twelve equation methods

| Experiment | Calibration method | Model parameters | | | Residual r.m.s. mm |
|---|---|---|---|---|---|
| | | $\Delta\rho_x$ mm | $\Delta\rho_x$ mm | $\Delta\rho_x$ mm | |
| Experiment #1 | *Six-equation* | 2.17 | 1.69 | -1.42 | 0.74 |
| | *Twelve-equation* | 2.07 | 1.66 | -1.30 | 0.75 |
| Experiment #2 | *Six-equation* | -0.53 | 0.59 | -1.76 | 0.20 |
| | *Twelve-equation* | -0.52 | 0.55 | -1.69 | 0.21 |
| Experiment #3 | *Six-equation* | 0.07 | 0.14 | 0.00 | 0.20 |
| | *Twelve-equation* | 0.12 | 0.00 | 0.10 | 0.21 |